\documentclass[letterpaper]{article} %
\usepackage{aaai24}  %
\usepackage{times}  %
\usepackage{helvet}  %
\usepackage{courier}  %
\usepackage[hyphens]{url}  %
\usepackage{graphicx} %
\def\UrlFont{\rm}  %
\usepackage{natbib}  %
\usepackage{caption} %
\usepackage{algorithm}
\usepackage[utf8]{inputenc} %
\usepackage[T1]{fontenc}    %
\usepackage{url}            %
\usepackage{booktabs}       %
\usepackage{amsfonts}       %
\usepackage{nicefrac}       %
\usepackage{microtype}      %
\usepackage{xcolor}         %
\usepackage{subcaption}
\usepackage{amsmath}
\usepackage{amssymb}
\usepackage{mathtools}
\usepackage{amsthm}
\usepackage{mathrsfs}
\usepackage{bm}
\usepackage{listings}
\usepackage{amsmath,bm}

\usepackage{xcolor} 
\newtheorem{thm}{Theorem}

\usepackage{multirow}

\providecommand{\R}{\mathbb{R}} %

\DeclareMathOperator{\expect}{\mathbb{E}}
\DeclareMathOperator{\E}{\expect}

\providecommand{\1}{\bm{1}}

\providecommand{\xx}{\bm{x}}

\providecommand{\muv}{\boldsymbol{\mu}}

\providecommand{\Xv}{\boldsymbol{X}}

\newcommand{\Var}{\mathrm{Var}}

\providecommand{\mX}{\bm{X}}

\newcommand{\ignore}[1]{}

\definecolor{color1}{RGB}{228,26,28}
\definecolor{color2}{RGB}{55,126,184}
\definecolor{color3}{RGB}{77,175,74}
\definecolor{color4}{RGB}{152,78,163}
\definecolor{color5}{RGB}{255,127,0}

\usepackage{pifont}
\newcommand{\cmark}{\ding{51}}%
\newcommand{\xmark}{\ding{55}}%

\makeatletter
\newcommand{\myitem}[1]{%
    \item[\textbf{(#1)}]\protected@edef\@currentlabel{#1}%
}
\makeatother

\def\rvm{{\mathbf{m}}}

\def\rvs{{\mathbf{s}}}
\def\rvt{{\mathbf{t}}}

\definecolor{DarkRed}{rgb}{0.5,0.1,0.1}

\newcommand{\highlightred}[1]{%
\colorbox{red!10}{$\displaystyle#1$}}

\newcommand{\highlightgreen}[1]{%
\colorbox{green!10}{$\displaystyle#1$}}
\newcommand{\highlightgreentext}[1]{%
  \colorbox{green!10}{#1}}

\newcommand{\highlightyellow}[1]{%
\colorbox{yellow!10}{$\displaystyle#1$}}
\newcommand{\highlightyellowtext}[1]{%
  \colorbox{yellow!10}{#1}}

\renewcommand{\epsilon}{\varepsilon}

\def\eqref#1{equation~\ref{#1}}

\def\1{\bm{1}}

\def\rvm{{\mathbf{m}}}

\def\rvs{{\mathbf{s}}}
\def\rvt{{\mathbf{t}}}

\def\vmu{{\bm{\mu}}}

\def\vsigma{{\bm{\sigma}}}

\def\vx{{\bm{x}}}

\def\mX{{\bm{X}}}

\DeclareMathAlphabet{\mathsfit}{\encodingdefault}{\sfdefault}{m}{sl}
\SetMathAlphabet{\mathsfit}{bold}{\encodingdefault}{\sfdefault}{bx}{n}

\def\emX{{X}}

\definecolor{codegreen}{rgb}{0,0.6,0}
\definecolor{codegray}{rgb}{0.5,0.5,0.5}
\definecolor{codepurple}{rgb}{0.58,0,0.82}
\definecolor{backcolour}{rgb}{0.95,0.95,0.92}

\lstdefinestyle{mystyle}{
    commentstyle=\color{codegreen},
    keywordstyle=\color{magenta},
    numberstyle=\tiny\color{codegray},
    stringstyle=\color{codepurple},
    basicstyle=\ttfamily\footnotesize,
    breakatwhitespace=false,         
    breaklines=true,                 
    captionpos=b,                    
    keepspaces=true,                 
    numbers=left,                    
    numbersep=5pt,                  
    showspaces=false,                
    showstringspaces=false,
    showtabs=false,                  
    tabsize=2
}

\usepackage{tikz}
\usetikzlibrary{arrows.meta, angles, quotes}
\definecolor{tabblue}{RGB}{31,119,180}
\definecolor{tabgreen}{RGB}{44,160,44}
\definecolor{tabred}{RGB}{214,39,40}

\usepackage{algpseudocodex}
\usepackage{todonotes}
\usepackage{enumitem}
\usepackage{makecell} %

\newcommand\extrafootertext[1]{%
    \bgroup
    \renewcommand\thefootnote{\fnsymbol{footnote}}%
    \renewcommand\thempfootnote{\fnsymbol{mpfootnote}}%
    \footnotetext[0]{#1}%
    \egroup
}

\usepackage{newfloat}
\usepackage{listings}
\DeclareCaptionStyle{ruled}{labelfont=normalfont,labelsep=colon,strut=off} %
\floatstyle{ruled}
\newfloat{listing}{tb}{lst}{}
\floatname{listing}{Listing}
\title{Ghost Noise for Regularizing Deep Neural Networks}
\author {
    Atli Kosson,
    Dongyang Fan,
    Martin Jaggi
}
\affiliations {
    EPFL, Switzerland\\
    firstname.lastname@epfl.ch
}

\begin{document}

\maketitle

\begin{abstract}
Batch Normalization (BN) is widely used to stabilize the optimization process and improve the test performance of deep neural networks.
The regularization effect of BN depends on the batch size and explicitly using smaller batch sizes with Batch Normalization, a method known as Ghost Batch Normalization (GBN), has been found to improve generalization in many settings.
We investigate the effectiveness of GBN by disentangling the induced ``Ghost Noise'' from normalization and quantitatively analyzing the distribution of noise as well as its impact on model performance.
Inspired by our analysis, we propose a new regularization technique called Ghost Noise Injection (GNI) that imitates the noise in GBN without incurring the detrimental train-test discrepancy effects of small batch training.
We experimentally show that GNI can provide a greater generalization benefit than GBN.
Ghost Noise Injection can also be beneficial in otherwise non-noisy settings such as layer-normalized networks, providing additional evidence of the usefulness of Ghost Noise in Batch Normalization as a regularizer.
\end{abstract}

\section{Introduction}
The adoption of normalization techniques has been a prevailing trend in the field of deep learning~\cite{ioffe2015batch, ba2016layer, wu2018group}, notably following the seminal introduction of Batch Normalization (BN) ~\cite{ioffe2015batch} in 2015.
Batch Normalization operates on each feature independently, normalizing it by subtracting its mean and dividing by its standard deviation.
The mean $\vmu$ and standard deviation (std) $\vsigma$ with elements corresponding to the different features in the input tensor.
During training, these statistics are computed over the batch dimension, causing the network output for a given sample to depend on the others in the batch.
This cross-sample dependency is undesirable at inference time, as data can arrive in small correlated batches or even one sample at a time.
Consequently, we utilize population-level statistics during inference, instead of computing the mean and variance for individual batches.
The population statistics are typically approximated by an exponential moving average of the mean and variance values observed during the training phase.

\begin{figure}[t]
\begin{center}
\includegraphics[width=1.0\columnwidth]{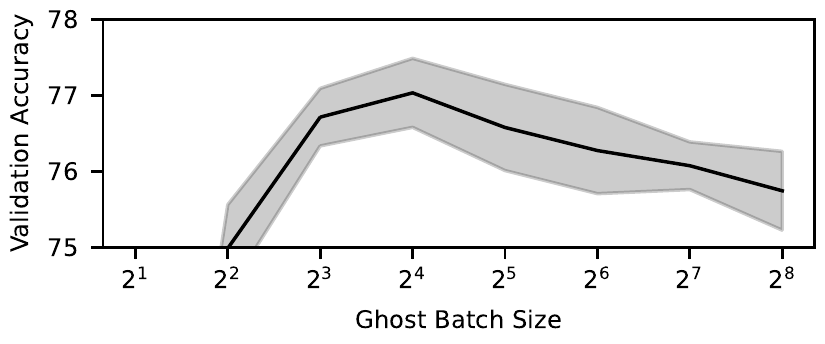}
\end{center}
\caption{Impact of ghost batch size on ResNet-18 accuracy on CIFAR-100. Mean (solid) and std (shaded) for 3 runs.}\label{fig:intro_gbn}
\end{figure}

The use of different forms of statistics in training and inference can lead to a \emph{detrimental} train-test discrepancy resulting in lower test performance.
One major difference is that during training each sample contributes to its normalization statistics, biasing the result.
The smaller the batch size, the greater the impact of each sample on the statistics, thus a larger bias.
We refer to this effect as the \textbf{self-dependency} of batch normalization.
The population statistics used during inference are identical for all samples and therefore not biased in the same way.
The distribution of normalized activations during testing can therefore be significantly different than during training, leading to worse generalization performance. This effect is studied by \citet{singh2019evalnorm} and \citet{summers2020things} which propose modifying inference to reduce the train-test discrepancy.
The use of correlated samples in a batch can similarly lead to a bias causing a different form of train-test discrepancy \cite{wu2021rethinking}.

Several alternative methods such as Layer Normalization~\cite{ba2016layer}, Instance Normalization~\cite{huang2017arbitrary}, Group Normalization~\cite{wu2018group}, Online Normalization~\cite{chiley2019online}, Batch Renormalization~\cite{ioffe2017batch} and Weight Normalization~\cite{salimans2016weight} have been proposed to overcome some of the issues with Batch Normalization.
These methods either normalize across other dimensions of the tensor $\Xv$, apply normalization to the weights instead of activations, or make use of the history over prior batches to augment the effective size of the batch.

Aside from the train-test discrepancy, the dependency of the batch normalization statistics on the elements in the batch can also have a \emph{beneficial} regularization effect, which is frequently observed but somewhat poorly understood.
From a geometrical point of view, \citet{keskar2017largebatch} empirically found that training with small batches arrives at flatter minima, which they argue contributes to better generalization.
\citet{hoffer2017train} proposed \textbf{Ghost Batch Normalization (GBN)}, followed by observations where performing batch normalization over smaller subgroups of the batch could improve generalization.
These smaller batches were termed \emph{ghost batches} by them. 
\citet{summers2020things} also observed this effect and recommended the use of GBN.
The regularization effect arises from the noise in $\vmu,\vsigma$ computed over a batch (typically randomly sampled) compared to the corresponding statistics for the full dataset.
This noise, which we will refer to as \textbf{Ghost Noise}, arises from the \textbf{cross-sample dependency} and increases for smaller batches, potentially explaining the improved generalization with GBN.
Figure~\ref{fig:intro_gbn} shows an example of how the ghost batch size can affect performance.

In this work, we analyze the effects of ghost noise and train-test discrepancy in BN.
The strength of both effects increases with smaller batch sizes, creating \emph{a trade-off between the regularization from the ghost noise and the detrimental effect of train-test discrepancy}.
Section~\ref{methods and analysis} constitutes the main part of our analysis and proposed methods.
We first propose \textbf{Exclusive Batch Normalization (XBN)}. This method mitigates the train-test discrepancy in GBN by simply excluding a sample from the computation of its normalization statistics. This way, self-dependency is eliminated during training, as in inference time. However, it requires moderately sized batches for stable training.
Following that, we introduce a novel technique named \textbf{Ghost Noise Injection (GNI)}.
GNI decouples ghost noise from the normalization process, allowing increased regularization from ghost noise without actually performing normalization using small batches.
This method avoids the detrimental effects associated with small batches and can be extended to BN-free settings, such as Layer Normalization, where inherent noise is absent.

Experimentally, we find that GNI can deliver a stronger regularization effect than GBN, resulting in improved test performance across a wide range of training settings.
Our ablation study suggests that both the noise in $\vmu$ (shifting noise) and in $\vsigma$ (scaling noise) contribute to the regularization effect of GNI.
In Convolutional Neural Networks (CNNs), the noise magnitude can vary between channels and layers depending on their distribution, in particular, how much of their variance comes from the spatial extent of a single sample compared to the variance between samples.
This ``adaptivity'' may be important, as we find that simpler i.i.d.\ (independent and identically distributed) noise methods are unable to match the regularization effect of GNI.
This includes several dropout variants as well as i.i.d.\ noise based on our analysis of the ghost noise distribution. 

\paragraph{Summary of Contributions:}
\setlist[itemize]{itemsep=0pt, topsep=0pt}
\begin{itemize}

\item We examine the impact of smaller batch sizes in GBN, focusing on the potentially useful regularization effect from ghost noise and the harmful train-test discrepancy, offering insights into the trade-off between the two.

\item We introduce Exclusive Batch Normalization (XBN) to mitigate the train-test discrepancy. Our findings also reveal that inherent self-dependency contributes to stability during small-batch training.

\item By deconstructing GBN into two constituent components, we devise a novel method termed Ghost Noise Injection (GNI). This technique allows us to specifically investigate the noise effects associated with small batch sizes.

\item Extensive experimentation highlights the utility of GNI as an effective regularization method across diverse architectures and training scenarios. This underscores the significance of the intrinsic noise within BN.

\end{itemize}

\section{Ghost Batch Normalization}
In this section we provide an overview of Ghost Batch Normalization (GBN) from \citet{hoffer2017train}, followed by a novel modeling of GBN as a composition of GBN and BN, separating the normalization and noise effects.

\subsection{Overview}
\label{section-gbn}
In short, GBN is performed by applying Batch Normalization on disjoint subsets of a batch, i.e. ghost batches.
By intentionally operating on smaller batch sizes, GBN increases the stochasticity in the normalization statistics compared to standard batch normalization.
This has been found to give a beneficial regularization effect in certain settings, see e.g. Figure~\ref{fig:intro_gbn}.
Our primary goal in this work is to explore and isolate the effect of Ghost Batch Normalization.

A batch ${\mX \in \mathbb{R}^{B\times C \times H \times W}}$ can be divided into ${G=B/N}$ ghost batches ${\{\mX_1, \mX_2, ...\mX_{G}\}}$, of size $N \times C \times H \times W$ each.
The output of GBN is obtained by performing batch normalization on each ghost batch.
For ghost batch $\mX_g$ with elements $\emX_{n,c,h,w}$, this can be expressed as:
\begin{equation}
    \widetilde{\mX}_g = \frac{(\mX_g - \vmu_g)}{\sqrt{\vsigma_g^2 + \epsilon}} \label{eq:gbn}
\end{equation}
where ${\vmu_g, \vsigma^2_g \in \mathbb{R}^{1\times C \times 1 \times 1}}$, ${\vmu_g = \frac{1}{NHW} \sum_{n,h,w} \emX_{n,c,h,w}}$, ${\vsigma^2_g = \frac{1}{NHW} \sum_{n,h,w} (\emX_{n,c,h,w} -\vmu_g)^2}$ and we use broadcasted operations similar to PyTorch~\cite{paszke2019pytorch}.

Here we note that the term ``batch'' is highly overloaded, encompassing different batch sizes of interest. \emph{Accelerator batch size} (denoted $B$) is the number of samples each worker (e.g.\ GPU) uses during a single forward / backward pass through the network. \emph{Normalization batch size} or ghost batch size (denoted $N$) is the number of samples over which normalization statistics are calculated. %

\subsection{Modeling GBN as Double Normalization}
\label{double normalization}
In this section, we present a novel perspective on GBN aimed at separating noise effects from normalization effects.  The key insight lies in the fact that applying standard batch normalization before ghost batch normalization preserves the output of GBN. For a batch $\Xv$, where batch normalization is denoted as $\text{BN}$ and ghost batch normalization as $\text{GBN}$, this relationship can be expressed as:
\begin{equation}
    \highlightyellow{\text{GBN}(\Xv)} = \highlightred{\text{GBN}(}\highlightgreen{\text{BN}(\Xv)}\highlightred{)}
\end{equation}
which follows from the fact that normalization is invariant to affine transformations of the inputs.

This decomposition of GBN into two successive normalization operations lets us isolate the differences between standard batch normalization and ghost batch normalization.
Ignoring $\epsilon$, the double normalization can be formulated as:
\begin{align}
    \hat{\Xv} &:= \highlightgreen{\text{BN}}(\Xv) = \frac{\Xv-\muv}{\vsigma} =[\hat{\Xv}_1,..,\hat{\Xv}_G] \label{eq:bn_gbn_1}\\
    \widetilde{\Xv}_g &:= \highlightred{\text{GBN}}(\hat{\Xv})_g = \frac{\hat{\Xv}_g-\hat{\muv}_g}{\hat{\vsigma}_g} \label{eq:bn_gbn_2}
\end{align}
where we split the normalized batch into $G$ ghost batches and show the following GBN for the $g$-th subgroup. Following the equivalence of \eqref{eq:gbn} and \eqref{eq:bn_gbn_2}, we have:

\begin{equation}
    \frac{\mX_g - \vmu_g}{\vsigma_g}  = \frac{1}{\hat{\vsigma}_g}\left(\frac{\Xv_g -\muv}{\vsigma}-\hat{\muv}_g \right)
\end{equation}
We can now write the $\muv_g$ and $\vsigma_g$ of the original $\text{GBN}(\Xv)$ setup in Equation~\ref{eq:gbn} as:
\begin{equation}\label{eq:deviation_statistics}
    \muv_g = \muv +\vsigma \hat{\muv}_g, \qquad \vsigma_g = \vsigma \hat{\vsigma}_g
\end{equation}
\highlightyellowtext{GBN} would be equivalent to \highlightgreentext{BN} if ${\hat{\muv}_g=0}$ and ${\hat{\vsigma}_g=1}$.
This happens when the ghost batches are identical, but generally, they contain different samples resulting in random variations in $\hat{\muv}_g$ and $\hat{\vsigma}_g$.
These fluctuations induce noise in two components, additive (or shifting) noise from $\hat{\muv}_g$ and multiplicative (or scaling) noise from $\hat{\vsigma}_g$.
The increased train-test discrepancy of GBN arises from the dependency of $\hat{\muv}_g$ and $\hat{\vsigma}_g$ on a specific sample (self-dependency).

\section{Methods and Analysis}\label{methods and analysis}
Within this section, we introduce two innovative approaches designed to enhance the generalization performance of Deep Neural Networks (DNNs):
1) Exclusive Batch Normalization, which addresses the issue of train-test discrepancy from self-dependency;
2) Ghost Noise Injection, a method that enhances training-time stochasticity by incorporating noise from ghost batches. We provide a detailed and clear analysis of this technique.

\subsection{Exclusive Batch
Normalization (XBN)}
\label{section-XBN}
We first consider a straightforward technique in which all elements within a ghost batch are utilized except for the element under consideration, for the calculation of normalization statistics.
This aims to mitigate the excessive reliance on a single sample during training. 
We term this approach \textbf{Exclusive Batch Normalization (XBN)}.
Note that XBN applies a separate set of normalization statistics for each sample.
Consequently, the training-time normalization statistics become less dependent on the specific sample itself, aligning more closely with the conditions at test time.
The mean and variance of the $k$-th sample in the $g$-th ghost batch of XBN is computed as:
\begin{align}
\vmu_{g,k} &= \frac{1}{(N-1)HW} \sum_{n \ne k,h,w} \emX_{n,c,h,w} \\
\vsigma^2_{g,k} &= \frac{1}{(N-1)HW} \sum_{n \ne k,h,w} (\emX_{n,c,h,w} -\vmu_{g,k})^2
\end{align}
Importantly, XBN maintains a similar level of noise as GBN while reducing the discrepancy between the train and test time. 
However, eliminating the self-dependency of normalization comes with a significant drawback: it does not guarantee a bounded output range. We present a detailed analysis in Appendix~\ref{apendix - XBN}. The absence of an output range constraint can lead to training instability. We empirically observe this behavior for smaller batch sizes. Yet, when the batch size is sufficiently large to support stable training, we do observe an increase in test accuracy, as demonstrated in Figure~\ref{fig:rn18_c100_val_overview}. This increase is likely attributed to the reduced train-test discrepancy.

XBN offers a way to address the train-test discrepancy in GBN which would in theory also allow us to obtain increased ghost noise by just decreasing the ghost batch size.
However, the instability of XBN for smaller batch sizes prevents this so we seek an improved method that does not suffer from this drawback.
Despite this XBN remains a valuable baseline for performance comparison.

\begin{figure*}[tb]
\centering
\begin{lstlisting}[language=Python]
import torch
def ghost_noise_injection(X: torch.Tensor, N: int, eps: float=1e-3):
    B, C, _, _ = X.shape  # Shape: Batch (B), Channels (C), Height (H), Width (W)
    with torch.no_grad():
        batch_var, batch_mean = torch.var_mean(X, dim=(0,2,3), correction=0)
        ghost_var = torch.zeros(size=(B,C), device=X.device)
        ghost_mean = torch.zeros(size=(B,C), device=X.device)
        for idx in range(B):
            ghost_sample = torch.randint(high=B, size=(N,))
            ghost_var[idx], ghost_mean[idx] = torch.var_mean(
                X[ghost_sample], dim=[0,2,3], correction=0)
        shift = (ghost_mean - batch_mean)
        scale = torch.sqrt((ghost_var + eps)/(batch_var + eps))
    return (X - shift.view(B, C, 1, 1)) / scale.view(B, C, 1, 1)
\end{lstlisting}
\caption{
A minimal PyTorch implementation of Ghost Noise Injection for convolutional activations maps without any performance optimizations. A small epsilon value helps with the numerical stability of the computation (1e-3 in all experiments).
}
\label{fig:gni_code}
\end{figure*}

\subsection{Ghost Noise Injection (GNI)}\label{sec:gni}
Building upon our findings in Section~\ref{double normalization}, we can identify $\hat{\muv}_g$ and $\hat{\vsigma}_g$ in Equation~\ref{eq:deviation_statistics} as the source of the ghost noise and the train-test discrepancy arising from self-dependency.
We propose replacing the GBN in double normalization with:
\begin{equation}
   \widetilde{\Xv} = \frac{\hat{\Xv}-\hat{\rvm}}{\hat{\rvs}} \label{eq:double_norm_gni}
\end{equation}
where the elements of $\hat{\rvm},\hat{\rvs} \in \mathbb{R}^{N\times C \times 1 \times 1}$ for a given input sample are computed as the mean and standard deviation of a \textbf{randomly sampled} (with replacement) ghost batch from $\hat{\Xv}$.
This departs from the methodology of GBN, where computations are based on the corresponding ghost batch.
Since the ghost batch from which the normalization statistics are calculated is randomly sampled, the current sample itself is not necessarily included in the ghost batch.
The self-dependency is thus reduced during training time.
We refer to the size of this subset as the ghost batch size, similar to GBN.
The resulting $\rvm$ and $\rvs$ are treated as pure noise, and \emph{we do not perform backpropagation through their computation}.

Equation~\ref{eq:double_norm_gni} assumes the inputs $\hat{\Xv}$ are batch normalized.
In general we may want to apply noise in more settings e.g.\ where no normalization is present.
This leads us to propose a more general form ---- Ghost Noise Injection (GNI):
\begin{equation}
\label{general-GNI}
    \frac{\Xv-(\rvm-\muv)}{\rvs/\vsigma}
\end{equation}
where $\muv$ and $\vsigma$ are computed like in batch normalization, and $\rvm$ and $\rvs$ are mean and standard deviation from a randomly sampled ghost batch of the \emph{unnormalized} input $\Xv$. $\rvm-\muv$ is treated as \textbf{shift noise} and $\rvs/\vsigma$ is treated as \textbf{scale noise}.
In Appendix~\ref{derivation - eq9} we give a derivation of Equation~\ref{general-GNI} by ``undoing'' the effect of the first normalization in double normalization.
A minimal implementation of GNI is shown in Figure~\ref{fig:gni_code}.
GNI offers two key advantages over GBN:
\begin{itemize}
    \item Each sample is unlikely to contribute strongly to its own normalization statistics, reducing the train-test discrepancy for a given level of noise (similar to XBN).
    \item The choice of normalization batch size $N$ is not limited to being a divisor of the accelerator batch size.
\end{itemize}
GNI comes at a slight computation and memory overhead which we discuss further in Appendix~\ref{appx:limitations}.

\subsection{Analytical Form of Ghost Noise Injection}
We further offer an analytical version of GNI, by modeling the distribution of the shift and scale noises.
The estimated mean $\hat{\muv}_g$ and standard deviation $\hat{\vsigma}_g$ from the $g$-th ghost batch, can be interpreted as bootstrapped statistics derived from the empirical distribution $P_{\hat{\Xv}}$.
Following the normalization in Equation~\ref{eq:bn_gbn_1}, the variable $\hat{\Xv}$ exhibits a zero mean and unit variance.
Assuming that the individual elements of $\hat{\Xv}$ are normally distributed allows us to derive an analytical distribution for the mean and variance, giving us additional insights into the workings of GBN.
For this section we focus on the distribution for a single channel $c$ in the $g$-th ghost batch, which we denote $\hat{\mu}_{gc}$ and $\hat{\sigma}_{gc}$.

\subsubsection{Case 1: Fully Connected Layers}
\label{noise-distr-FNN}
Assuming that the output of BN is independent and normally distributed, the normalization statistics $\hat{\muv}_g$ and $\hat{\vsigma}_g^2$ are computed over a ghost batch $\hat{\Xv}_g=[\hat{\xx}_1,..,\hat{\xx}_N] \in \R^{N\times C}$ of $N$ elements sampled i.i.d. from $\mathcal{N}(0,1)$.
We can then derive the distribution of the sample mean, and therefore the shift noise, as:
\begin{equation}
\label{shift-noise-1d-distr}
     \hat{\mu}_{gc} = \frac{1}{N}\sum_{n=1}^N \hat{x}_{nc} \sim \mathcal{N}\left(0,\frac{1}{N}\right)
\end{equation}
The sum of the square of independent standard normally distributed variables follows a Chi-squared distribution:
\begin{equation}
\label{scale-noise-1d-distr}
     \hat{\sigma}_{gc}^2 = \frac{1}{N}\sum_{n=1}^N (\hat{x}_{nc}-0)^2 \sim \frac{1}{N}\mathcal{\chi}^2 (N)
\end{equation}

This clearly shows the dependency of the noise on the ghost batch size.
Larger ghost batch sizes correspond to less noise, explaining their reduced generalization benefit as observed in e.g.\ Figure~\ref{fig:intro_gbn}.

\begin{itemize}
    \item \textbf{Analytical Ghost Noise Injection:}
Based on the preceding analysis, an alternative approach to computing ghost statistics from sampled batches is to directly sample $\hat{\muv}_g$ and $\hat{\vsigma}_g^2$ from the respective analytical distributions for each channel. Let $\hat{\Xv}$ denote $\Xv$ post batch normalization, as previously defined. Subsequently, for each channel $c$, we can then inject noise using:
\begin{equation}
\label{NAT}
    \frac{\hat{\Xv_c}-\mu_c}{\sigma_c} \qquad \text{with}\: \mu_c \sim P_{\hat{\mu}_{gc}}, \sigma_c \sim P_{\hat{\sigma}_{gc}^2}
\end{equation}
$P_{\hat{\mu}_{gc}}$ is given in \eqref{shift-noise-1d-distr} and $ P_{\hat{\sigma}_{gc}^2}$ is given in \eqref{scale-noise-1d-distr}. The hyperparameter $N$ is used to vary the amount of noise, corresponding to different ghost batch sizes. 
\item \textbf{Comparison to Gaussian Dropout:}
 It is noteworthy to highlight the  similarity between the scaling noise and Gaussian Dropout, which is written as:
\begin{equation}
    \textstyle
    \hat{\Xv} \cdot \rvt \qquad \text{with}\: \rvt \sim \mathcal{N}(1,\frac{p}{1-p}) %
\end{equation}
where $\rvt$ is sampled element-wise from the Gaussian distribution and $p$ is interpreted like the drop probability in standard dropout ($p=0$ for no dropout).
Both the scaling noise and Gaussian dropout are multiplicative with similar yet slightly different distributions.
\end{itemize}

\subsubsection{Case 2: Convolutional Layers}
\label{noise distr - Conv}
In CNNs the batch statistics are computed across both the batch and spatial dimensions (e.g.\ the height and width of an image).
As a result, the data can exhibit variability in two distinct manners: variation among samples along the batch dimension, termed \emph{inter-sample variance}, and variation within a single sample across the spatial dimension, referred to as \emph{intra-sample variance}.
This stands in contrast to the fully connected scenario, where all variance emerges along the batch dimension, and the concept of intra-sample variance is absent.

\begin{figure*}[t]
    \centering
    \includegraphics[width=0.9\textwidth]{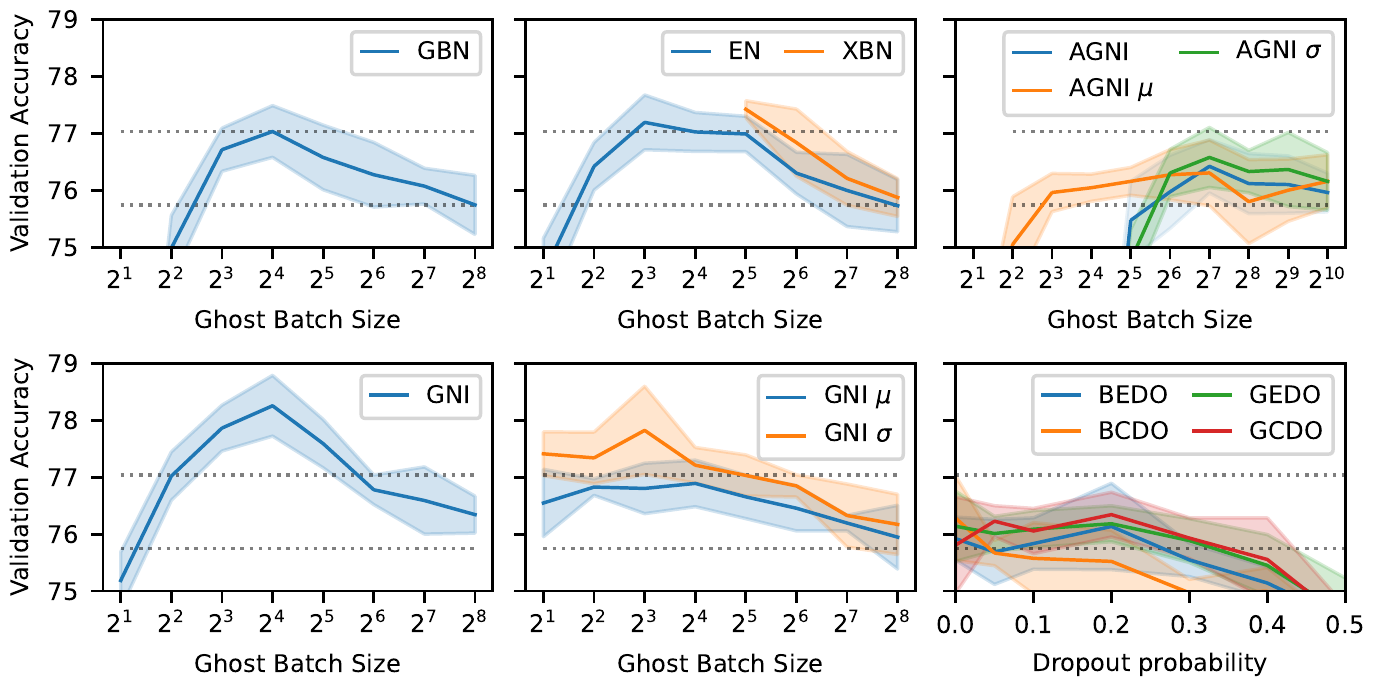}
    \caption{
    CIFAR-100 ResNet-18 validation accuracy versus ghost batch size and dropout probability for different methods.
    Each line is the average of five runs and the shaded area shows the standard deviation.
    The dotted lines show the standard batch normalization performance and the maximum for ghost batch normalization.
    For dropout (DO, lower right panel), B=Bernoulli, G=Gaussian, E=elementwise and C=channelwise. See Experiments section for further details and discussion.
    }
    \label{fig:rn18_c100_val_overview}
\end{figure*}

To gain insight into how the inter- and intra-sample variances could affect the noise distributions we will present and analyze a simplified model.
We will assume that ${\Xv \in \mathbb{R}^{B\times C\times I}}$, where $I$ is the spatial dimension e.g.\ ${I=H\times W}$ where $H$ is the height and $W$ the width for image data.
We further assume the true mean of $b$-th batch in a specific channel $c$ is sampled i.i.d. from a normal distribution, i.e. $\mu_{bc} \overset{i.i.d.}{\sim} \mathcal{N}(0,\sigma_{Bc}^2)$, where $\sigma_{Bc}^2$ denotes the inter-sample variance and only varies across $c$, i.e.\ it is constant across $B$. %
We model the $i$-th spatial location in the $b$-th batch as being sampled from ${x_{bci} \overset{i.i.d.}{\sim} \mathcal{N}(\mu_{bc},\sigma_{Ic}^2)}$, with an intra-sample variance $\sigma_{Ic}^2$ that does not vary across $I$.
After batch normalization, we always have unit variance, which means $\sigma_{Bc}^2+\sigma_{Ic}^2 =1$, for any $c$. Now assume we sample a random  ghost batch of size $N$. As the sample mean is an average of all samples, we still have it following a normal distribution.
We defer the calculation details of the distributions to Appendix~\ref{appendix - noise distr}.
The resulting expressions for the distributions of the scale and shift noise in channel $c$ are:

\begin{equation}
  \hat{\mu}_{gc} \sim \mathcal{N}\left(0,\frac{1}{NI}\sigma_{Ic}^2 +\frac{1}{N}\sigma_{Bc}^2\right)
\end{equation}

\begin{equation}
\label{scale-noise-distr-2d}
  \hat{\sigma}_{gc}^2 \sim \frac{\sigma_{Ic}^2}{NI}\chi^2(NI) + \frac{\sigma_{Bc}^2}{N}\chi^2(N)
\end{equation}

When $\sigma_{Ic}^2=0$, the outcomes align with the 1D scenario. However, the analysis in the 2D case affords us greater flexibility to distinguish between inter-sample variance ($\sigma_{Bc}^2$) and intra-sample variance ($\sigma_{Ic}^2$). An additional noteworthy insight pertains to channel dependency.
Given that both $\sigma_{Bc}^2$ and $\sigma_{Ic}^2$ are specific to individual channels, a distinct noise effect is introduced for each channel.
This phenomenon will be empirically explored in Figure~\ref{fig:distributions}.

\begin{table}[b]
\centering
\begin{tabular}{lccccc}
    \toprule
     & BN    & GBN-16 & GNI-16 & XBN   & EN    \\
    \midrule
    Mean & 77.10 & 78.20 & \textbf{78.84} & 78.33 & 78.33 \\
    Std  & 0.26 & 0.10 & 0.09 & 0.40 & 0.27 \\
    \bottomrule
\end{tabular}
\caption{CIFAR-100 test accuracy (mean$\pm$std\% for 3 runs).}\label{table:Cifar100-ResNet18}
\end{table}

\section{Experiments and Discussions}\label{sec:experiments}
The setup of our experiments is detailed in Appendix~\ref{appx:experimental_details}. 
\subsection{Comparison of Methods}
\paragraph{Baseline - GBN} 
The top left panel of Figure~\ref{fig:rn18_c100_val_overview} shows how the ghost batch size affects the final validation accuracy when using Ghost Batch Normalization.
Using a ghost batch size $N=256$ is equivalent to standard batch normalization.
We see that the accuracy increases for smaller batch sizes up to a certain extent, after which it goes down again.
The initial improvement could be attributed to heightened regularization, while the subsequent decrease might arise from excessive regularization or the accentuated bias in normalization statistics, potentially leading to a discrepancy between training and testing.
The optimal $N=16$ gives an accuracy boost of just over 1\% on both the validation and the test sets (Table~\ref{table:Cifar100-ResNet18}).

\begin{figure*}[tb]
    \centering
    \includegraphics[width=0.9\textwidth]{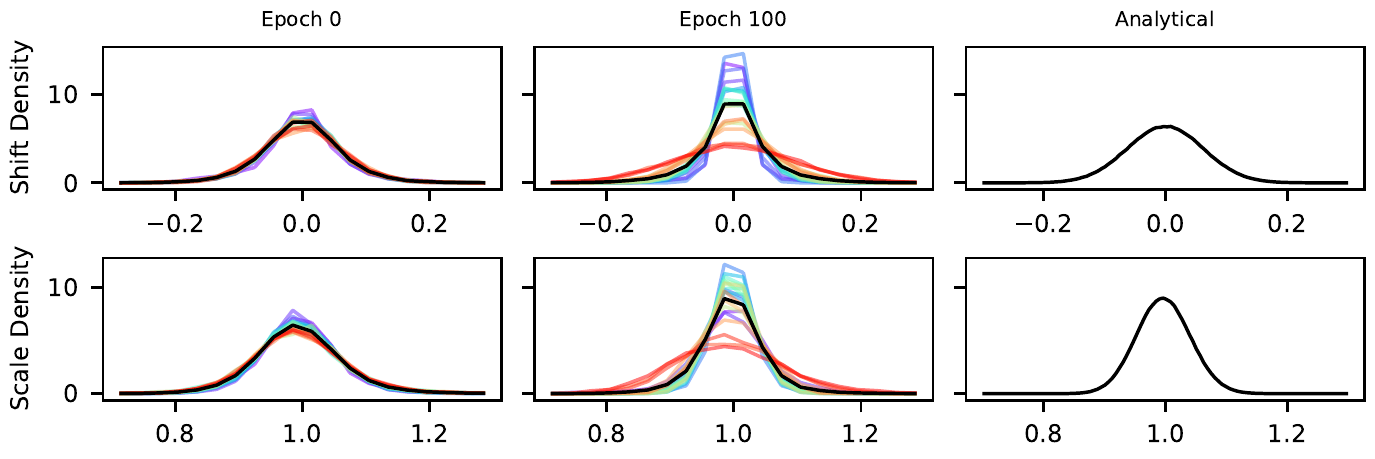}
    \caption{Measured GNI noise distributions for ResNet18 on CIFAR100 with a Ghost Batch Size of 32 along with analytical fully connected distributions for batch size 256.
    Each GNI line is an average over a layer, the distributions also vary between channels inside each layer.
    The lines are plotted with a standard color spectrum (rainbow) from violet (first layer) to red (last layer).
    }
    \label{fig:distributions}
\end{figure*}

\paragraph{XBN}
The upper-middle panel of Figure~\ref{fig:rn18_c100_val_overview} delves into the efficacy assessment of Exclusive Batch Normalization (XBN) and EvalNorm~\citep{singh2019evalnorm} (EN). 
The primary objective of both techniques is the enhancement of test performance through the mitigation of the train-test discrepancy. XBN accomplishes this by altering sample interdependence during training to align more closely with inference conditions, while EvalNorm adopts the reverse approach.
XBN can be unstable for small batch sizes (as discussed in Section~\ref{section-XBN}), in this case batch sizes smaller than 32.
Despite this, both XBN and EN appear to effectively alleviate certain aspects of the train-test discrepancy, yielding a marginal accuracy increase in comparison to GBN.
Specifically, XBN exhibits slightly superior performance on the validation set, while producing comparable outcomes on the test set (as indicated in Table~\ref{table:Cifar100-ResNet18}).

\paragraph{Sample-based GNI}
Within Figure~\ref{fig:rn18_c100_val_overview}, the lower-left panel presents the performance evaluation of Ghost Noise Injection (GNI) across varying ghost batch sizes. Remarkably, we note a substantial enhancement—approximately double that of GBN. 
At higher ghost batch sizes, GNI performs similarly to XBN.
This observation suggests that GNI effectively mitigates the train-test discrepancy while stabilizing training with smaller batch sizes, allowing us to benefit from their increased regularization effect.
Notably, among the explored methodologies, GNI emerges as the most successful, showcasing superior performance on the test set (Table~\ref{table:Cifar100-ResNet18}.
In the lower middle panel of Figure~\ref{fig:rn18_c100_val_overview} we investigate the effect of the different noise components of GNI.
GNI~$\muv$ only includes the shift and GNI~$\vsigma$ only the scaling term. Evidently, each component independently delivers a significant performance enhancement. 
However, their individual impacts fall short of matching the comprehensive benefits of GNI.
This strongly implies that both constituents of Ghost Noise collectively yield advantageous outcomes within this context.

\paragraph{Ghost Noise Distribution} 
Figure~\ref{fig:distributions} shows measured distributions of both the scale and shift components of the noise generated by GNI. 
The observed average distributions are similar to our derived distribution for the fully connected case, but the batch size parameter must be adjusted to account for the intra-sample variance component, as elaborated in \eqref{scale-noise-distr-2d}. 
The depicted distributions exhibit notable variations across layers and channels (though not visually presented here), particularly in the later stages of training. This variability could potentially be attributed to fluctuations in the intra-sample variance.
In the top right panel of Figure~\ref{fig:rn18_c100_val_overview} we apply our analytical GNI \textbf{(AGNI)} to training.
Although we observe some improvements, it is unable to match the performance of the sample-based GNI. This suggests a potential significance associated with the \emph{channel-specific} fluctuations.
In Appendix~\ref{appx:more_experiments} we further measure the distribution of ghost noise for a fully connected network.

\paragraph{GNI vs Dropout}
GNI is a regularizer and a potential alternative to Dropout.
In Figure~\ref{fig:rn18_c100_val_overview} (bottom right) we compare four variants of dropout (Bernoulli/Gaussian and Element-wise/Channel-wise) to the other methods.
We apply the dropout after the second normalization layer on each branch, as was done in Wide Residual Networks~\cite{zagoruyko2017wide}.
We find that dropout performs similarly to AGNI but is unable to match GBN or sample-based GNI.

\begin{figure}[htb]
\begin{center}
\includegraphics[width=1.0\columnwidth]{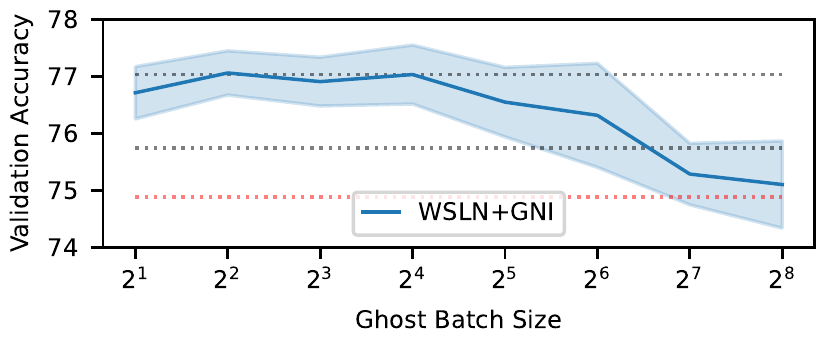}
\end{center}
\caption{Accuracy impact of Ghost Noise Injection on Weight Standardized Layer Normalized (WSLN) ResNet18 trained on CIFAR100. The two upper black dotted lines correspond to Figure~\ref{fig:rn18_c100_val_overview} (BN baseline, best GBN performance) and the bottom red dotted line to WSLN without GNI.}\label{fig:rn18_ln_val}
\end{figure}
\subsection{Wider Applicabilities of GNI}
\paragraph{Applicability to other Normalizers}
So far we have applied GNI on top of Batch Normalization.
However, GNI is not limited to this use case.
Figure~\ref{fig:rn18_ln_val} shows the CIFAR-100 validation accuracy of a ResNet-18 where batch normalization has been replaced by weight-standardization~\cite{qiao2020WS} and layer normalization~\cite{ba2016layer}.
We find that GNI can also provide a significant accuracy boost in this setting, bringing the performance with this noise-free normalization method in line with that of GBN.

\paragraph{Generalization to other Datasets}
Table~\ref{tab:more_datasets} shows that GNI can also regularize the training of ResNet-50 on ImageNet-1k and ResNet-20 on CIFAR-10.
In both cases, GNI provides a decent boost in accuracy and outperforms GBN.

\begin{table}[bt]
    \centering
    \begin{tabular}{lcc}
        \toprule
        Setup & CIFAR-10 RN-20 & ImageNet-1k RN-50 \\
        \midrule
        BN-256  & 92.22$\pm$0.05 & 77.43 \\
        GBN-16  & 92.65$\pm$0.07 & 76.90 \\
        GBN-32  & 92.53$\pm$0.33 & 77.14 \\
        GBN-64  & 92.50$\pm$0.13 & 77.48 \\
        GNI-16  & \textbf{93.01}$\pm$0.28 & 77.62 \\
        GNI-32  & 92.76$\pm$0.11 & 77.70 \\
        GNI-64  & 92.56$\pm$0.12 & \textbf{77.98} \\
        \bottomrule
    \end{tabular}
    \caption{Test accuracy for ResNet-20 CIFAR-10 (mean$\pm$std\% for 3 runs) and ResNet-50 ImageNet-1k for different normalization and noise setups. GNI is performed on top of the standard BN-256 for additional regularization.}\label{tab:more_datasets}
\end{table}

\paragraph{Extension to other Architectures} To investigate the adaptability of GNI to diverse architectures, we perform additional experiments on CIFAR-10 using a Normalization Free ResNet~\cite{brock2021high}, a Simple Visual Transformer~\cite{beyer2022better,dosovitskiy2021image,Cordonnier2020On} and a ConvMixer~\cite{trockman2023patches}.
We explore two data augmentation settings, no augmentation and the ResNet style augmentation which consists of a random flip, padding and a random crop.
Further details about the networks and other experimental details are given in Appendix~\ref{appx:experimental_details}.
In each case, GNI can improve the test accuracy showing that it can generalize to other architectures.
We typically observe a greater performance boost when weaker data augmentation is used.
Excessive data augmentation may already prevent a network of a given size that is trained for limited time from fitting the training data, reducing the benefit of additional regularization.
The ghost batch size $N$ was tuned on a validation set and varies considerably between the settings.

\begin{table}[t]
    \centering
    \label{table-3}
    \begin{tabular}{lcccc}
        \toprule
        Network & Aug & Baseline & GNI (N) \\
        \midrule
        NF ResNet & \xmark & 84.31$\pm$0.86 & 86.18$\pm$0.58~(\hphantom{0}32) \\
        NF ResNet & \cmark & 93.60$\pm$0.21 & 94.19$\pm$0.24~(\hphantom{0}32) \\
        Simple ViT & \xmark & 71.45$\pm$0.12 & 74.36$\pm$0.27~(\hphantom{00}4) \\
        Simple ViT & \cmark & 88.19$\pm$0.08 & 89.55$\pm$0.05~(\hphantom{0}12) \\
        ConvMixer & \xmark & 91.18$\pm$0.24 & 91.57$\pm$0.08~(192) \\
        ConvMixer & \cmark & 93.29$\pm$0.17 & 93.66$\pm$0.12~(128) \\
        \bottomrule
    \end{tabular}
    \caption{Test Accuracy (mean$\pm$std\% for 3 runs) on CIFAR-10 with/without data augmentation (Aug) and with/without GNI (using ghost batch size N). See Appendix~\ref{appx:experimental_details} for details.}
\end{table}

\section{Related Work}
\label{sec:related work}

\paragraph{Mitigating Train-test Discrepancy}  \citet{singh2019evalnorm} proposed EvalNorm to  estimate corrected normalization
statistics to use for BN during evaluation, such that training and testing time normalized activation distributions become closer. \citet{summers2020things} incorporated example statistics during inference, which reduces the output range of a network with Batch Normalization during testing. While both works try to mitigate train-test discrepancy by alternating test time normalization statistics, XBN shows that it is also possible to address the same issue by reducing sample dependency during training time.

\paragraph{Noise Injection}
\citet{liu2022nomorelization} proposed a composition of two trainable scalars and a zero-centered noise injector for regularizing normalization-free DNNs.
\citet{camuto2020explicit} analyzed Gaussian Noise Injection from a theoretical point of view and found injected noise particularly regularises neural network layers closer to the output.
We compare the GNI to these methods in Appendix~\ref{appx:more_experiments}.
Compared to these two, Ghost Noise Injection more closely imitates the noise of batch normalization, accounting for the channel and layer-wise differences in the distribution.
\citet{shekhovtsov2018stochastic} explored the noise in batch normalization from a Bayesian perspective, obtaining similar results to Equations~\ref{shift-noise-1d-distr}-\ref{scale-noise-1d-distr}, but did not analyze the layer and channel dependencies.

\paragraph{Dropout}  Dropout was first proposed by ~\citet{JMLR:v15:srivastava14a} as a simple regularization approach to prevent units from co-adapting too much. \citet{wei2020implicit} demonstrates the explicit and implicit regularization effects from dropout and found out that the implicit regularization effect is analogous to the effect of stochasticity in small mini-batch stochastic gradient descent. \citet{c-dropout-2-hou, C-dropout_He_2022} applied channel-wise dropout and experimentally showed consistent benefits in DNNs with convolutional structures. Further, it is observed that channel-wise dropout encourages its succeeding layers to minimize the intra-class feature variance~\cite{C-dropout_He_2022}.

\section{Conclusion}

In this study, we explored an important aspect of batch normalization -- the generalization benefit from smaller batch sizes.
By formulating GBN as a series of two normalization operations, we are able to analyze the impact of smaller batch sizes on the noise and train-test discrepancy.
Our analysis of the noise component unveiled its channel-dependent nature, comprising two distinct facets—shifting and scaling—both instrumental in the overall effectiveness.
Furthermore, we demonstrated that the train-test discrepancy could be alleviated by preventing a sample from contributing to its own normalization statistics during training.
Building on these insights, we introduced Ghost Noise Injection (GNI), a novel technique that elevates ghost noise levels without necessitating normalization over smaller batches, thus diminishing the train-test mismatch. 
Empirical investigations across various scenarios, including those without batch normalization, showcased GNI's beneficial impact on generalization, underscoring ghost noise's significance as a potent source of regularization in batch normalization.

\section*{Acknowledgments}
We thank Maksym Andriushchenko for his feedback on the manuscript, which has enhanced its quality and clarity.

\bibliography{aaai24}

\begin{thebibliography}{32}
\providecommand{\natexlab}[1]{#1}

\bibitem[{Ba, Kiros, and Hinton(2016)}]{ba2016layer}
Ba, J.~L.; Kiros, J.~R.; and Hinton, G.~E. 2016.
\newblock Layer normalization.
\newblock \emph{arXiv preprint arXiv:1607.06450}.

\bibitem[{Beyer, Zhai, and Kolesnikov(2022)}]{beyer2022better}
Beyer, L.; Zhai, X.; and Kolesnikov, A. 2022.
\newblock Better plain ViT baselines for ImageNet-1k.
\newblock \emph{arXiv preprint arXiv:2205.01580}.

\bibitem[{Brock et~al.(2021)Brock, De, Smith, and Simonyan}]{brock2021high}
Brock, A.; De, S.; Smith, S.~L.; and Simonyan, K. 2021.
\newblock High-performance large-scale image recognition without normalization.
\newblock In \emph{International Conference on Machine Learning}, 1059--1071. PMLR.

\bibitem[{Camuto et~al.(2020)Camuto, Willetts, Simsekli, Roberts, and Holmes}]{camuto2020explicit}
Camuto, A.; Willetts, M.; Simsekli, U.; Roberts, S.~J.; and Holmes, C.~C. 2020.
\newblock Explicit regularisation in gaussian noise injections.
\newblock \emph{Advances in Neural Information Processing Systems}, 33: 16603--16614.

\bibitem[{Chiley et~al.(2019)Chiley, Sharapov, Kosson, Koster, Reece, Samaniego de~la Fuente, Subbiah, and James}]{chiley2019online}
Chiley, V.; Sharapov, I.; Kosson, A.; Koster, U.; Reece, R.; Samaniego de~la Fuente, S.; Subbiah, V.; and James, M. 2019.
\newblock Online normalization for training neural networks.
\newblock \emph{Advances in Neural Information Processing Systems}, 32.
\newblock ArXiv:1905.05894.

\bibitem[{Cordonnier, Loukas, and Jaggi(2020)}]{Cordonnier2020On}
Cordonnier, J.-B.; Loukas, A.; and Jaggi, M. 2020.
\newblock On the Relationship between Self-Attention and Convolutional Layers.
\newblock In \emph{International Conference on Learning Representations}.

\bibitem[{Deng(2012)}]{deng2012mnist}
Deng, L. 2012.
\newblock The mnist database of handwritten digit images for machine learning research.
\newblock \emph{IEEE Signal Processing Magazine}, 29(6): 141--142.

\bibitem[{Dosovitskiy et~al.(2021)Dosovitskiy, Beyer, Kolesnikov, Weissenborn, Zhai, Unterthiner, Dehghani, Minderer, Heigold, Gelly, Uszkoreit, and Houlsby}]{dosovitskiy2021image}
Dosovitskiy, A.; Beyer, L.; Kolesnikov, A.; Weissenborn, D.; Zhai, X.; Unterthiner, T.; Dehghani, M.; Minderer, M.; Heigold, G.; Gelly, S.; Uszkoreit, J.; and Houlsby, N. 2021.
\newblock An Image is Worth 16x16 Words: Transformers for Image Recognition at Scale.
\newblock arXiv:2010.11929.

\bibitem[{He et~al.(2016)He, Zhang, Ren, and Sun}]{he2016deep}
He, K.; Zhang, X.; Ren, S.; and Sun, J. 2016.
\newblock Deep residual learning for image recognition.
\newblock In \emph{Proceedings of the IEEE conference on computer vision and pattern recognition}, 770--778.

\bibitem[{He et~al.(2022)He, Zhang, Shan, Liu, Wu, and Chen}]{C-dropout_He_2022}
He, M.; Zhang, J.; Shan, S.; Liu, X.; Wu, Z.; and Chen, X. 2022.
\newblock Locality-Aware Channel-Wise Dropout for Occluded Face Recognition.
\newblock \emph{{IEEE} Transactions on Image Processing}, 31: 788--798.

\bibitem[{Hoffer, Hubara, and Soudry(2017)}]{hoffer2017train}
Hoffer, E.; Hubara, I.; and Soudry, D. 2017.
\newblock Train longer, generalize better: closing the generalization gap in large batch training of neural networks.
\newblock \emph{Advances in neural information processing systems}, 30.

\bibitem[{Hou and Wang(2019)}]{c-dropout-2-hou}
Hou, S.; and Wang, Z. 2019.
\newblock Weighted Channel Dropout for Regularization of Deep Convolutional Neural Network.
\newblock In \emph{Proceedings of the Thirty-Third AAAI Conference on Artificial Intelligence and Thirty-First Innovative Applications of Artificial Intelligence Conference and Ninth AAAI Symposium on Educational Advances in Artificial Intelligence}, AAAI'19/IAAI'19/EAAI'19. AAAI Press.
\newblock ISBN 978-1-57735-809-1.

\bibitem[{Huang and Belongie(2017)}]{huang2017arbitrary}
Huang, X.; and Belongie, S. 2017.
\newblock Arbitrary style transfer in real-time with adaptive instance normalization.
\newblock In \emph{Proceedings of the IEEE international conference on computer vision}, 1501--1510.

\bibitem[{Ioffe(2017)}]{ioffe2017batch}
Ioffe, S. 2017.
\newblock Batch renormalization: Towards reducing minibatch dependence in batch-normalized models.
\newblock \emph{Advances in neural information processing systems}, 30.

\bibitem[{Ioffe and Szegedy(2015)}]{ioffe2015batch}
Ioffe, S.; and Szegedy, C. 2015.
\newblock Batch normalization: Accelerating deep network training by reducing internal covariate shift.
\newblock In \emph{International conference on machine learning}, 448--456. pmlr.

\bibitem[{Keskar et~al.(2017)Keskar, Mudigere, Nocedal, Smelyanskiy, and Tang}]{keskar2017largebatch}
Keskar, N.~S.; Mudigere, D.; Nocedal, J.; Smelyanskiy, M.; and Tang, P. T.~P. 2017.
\newblock On Large-Batch Training for Deep Learning: Generalization Gap and Sharp Minima.
\newblock arXiv:1609.04836.

\bibitem[{Krizhevsky and Hinton(2009)}]{cifar100}
Krizhevsky, A.; and Hinton, G. 2009.
\newblock CIFAR-100 (Canadian Institute For Advanced Research).
\newblock \emph{Dataset available from https://www.cs.toronto.edu/~kriz/cifar.html}.

\bibitem[{Liu et~al.(2022)Liu, Yang, Ding, and Lu}]{liu2022nomorelization}
Liu, C.; Yang, Y.; Ding, Y.; and Lu, H. 2022.
\newblock NoMorelization: Building Normalizer-Free Models from a Sample's Perspective.
\newblock arXiv:2210.06932.

\bibitem[{Paszke et~al.(2019)Paszke, Gross, Massa, Lerer, Bradbury, Chanan, Killeen, Lin, Gimelshein, Antiga et~al.}]{paszke2019pytorch}
Paszke, A.; Gross, S.; Massa, F.; Lerer, A.; Bradbury, J.; Chanan, G.; Killeen, T.; Lin, Z.; Gimelshein, N.; Antiga, L.; et~al. 2019.
\newblock Pytorch: An imperative style, high-performance deep learning library.
\newblock \emph{Advances in neural information processing systems}, 32.

\bibitem[{Qiao et~al.(2020)Qiao, Wang, Liu, Shen, and Yuille}]{qiao2020WS}
Qiao, S.; Wang, H.; Liu, C.; Shen, W.; and Yuille, A. 2020.
\newblock Micro-Batch Training with Batch-Channel Normalization and Weight Standardization.
\newblock arXiv:1903.10520.

\bibitem[{Salimans and Kingma(2016)}]{salimans2016weight}
Salimans, T.; and Kingma, D.~P. 2016.
\newblock Weight normalization: A simple reparameterization to accelerate training of deep neural networks.
\newblock \emph{Advances in neural information processing systems}, 29.

\bibitem[{Shekhovtsov and Flach(2019)}]{shekhovtsov2018stochastic}
Shekhovtsov, A.; and Flach, B. 2019.
\newblock Stochastic normalizations as bayesian learning.
\newblock In \emph{Computer Vision--ACCV 2018: 14th Asian Conference on Computer Vision, Perth, Australia, December 2--6, 2018, Revised Selected Papers, Part II 14}, 463--479. Springer.

\bibitem[{Singh and Shrivastava(2019)}]{singh2019evalnorm}
Singh, S.; and Shrivastava, A. 2019.
\newblock Evalnorm: Estimating batch normalization statistics for evaluation.
\newblock In \emph{Proceedings of the IEEE/CVF International Conference on Computer Vision}, 3633--3641.

\bibitem[{Srivastava et~al.(2014)Srivastava, Hinton, Krizhevsky, Sutskever, and Salakhutdinov}]{JMLR:v15:srivastava14a}
Srivastava, N.; Hinton, G.; Krizhevsky, A.; Sutskever, I.; and Salakhutdinov, R. 2014.
\newblock Dropout: A Simple Way to Prevent Neural Networks from Overfitting.
\newblock \emph{Journal of Machine Learning Research}, 15(56): 1929--1958.

\bibitem[{Summers and Dinneen(2020)}]{summers2020things}
Summers, C.; and Dinneen, M.~J. 2020.
\newblock Four Things Everyone Should Know to Improve Batch Normalization.
\newblock arXiv:1906.03548.

\bibitem[{Trockman and Kolter(2023)}]{trockman2023patches}
Trockman, A.; and Kolter, J.~Z. 2023.
\newblock Patches Are All You Need?
\newblock \emph{Transactions on Machine Learning Research}.
\newblock Featured Certification.

\bibitem[{Wang(2023)}]{vitpytorch}
Wang, P. 2023.
\newblock vit-pytorch.
\newblock \url{https://github.com/lucidrains/vit-pytorch}.

\bibitem[{Wei, Kakade, and Ma(2020)}]{wei2020implicit}
Wei, C.; Kakade, S.; and Ma, T. 2020.
\newblock The implicit and explicit regularization effects of dropout.
\newblock In \emph{International conference on machine learning}, 10181--10192. PMLR.

\bibitem[{Wightman(2019)}]{rw2019timm}
Wightman, R. 2019.
\newblock PyTorch Image Models.
\newblock \url{https://github.com/rwightman/pytorch-image-models}.

\bibitem[{Wu and He(2018)}]{wu2018group}
Wu, Y.; and He, K. 2018.
\newblock Group normalization.
\newblock In \emph{Proceedings of the European conference on computer vision (ECCV)}, 3--19.

\bibitem[{Wu and Johnson(2021)}]{wu2021rethinking}
Wu, Y.; and Johnson, J. 2021.
\newblock Rethinking" batch" in batchnorm.
\newblock \emph{arXiv preprint arXiv:2105.07576}.

\bibitem[{Zagoruyko and Komodakis(2017)}]{zagoruyko2017wide}
Zagoruyko, S.; and Komodakis, N. 2017.
\newblock Wide Residual Networks.
\newblock arXiv:1605.07146.

\end{thebibliography}
\newpage
\appendix
\onecolumn

\section{Experimental Details}\label{appx:experimental_details}
Our experimental codes are written in PyTorch~\cite{paszke2019pytorch} and based on the TIMM~\cite{rw2019timm} library.

\paragraph{Hyperparameter Tuning}
The optimization hyperparameters employed in each experiment are grounded in pre-existing hyperparameters from the original works. 
We generally perform a light tuning of the baseline learning rate (all CIFAR-based experiments) and the warmup period (for extension to other architectures).
This tuning is performed on a validation set created by randomly sampling 10\% of the original training set.
For Ghost Noise Injection we sweep over batch sizes in the set $\{2^1,2^2,\ldots,2^8\}$, except for extension to other architectures where we also consider the halfway points between these.
This tuning is also performed on a validation set and typically averaged over three seeds.
We do not re-tune other hyperparameters (e.g. the learning rate) for use with GNI.

\subsection{ResNet Experiments}
\paragraph{Base Setup} 
We employ ResNet-18~\cite{he2016deep} for training on the CIFAR-100 dataset~\cite{cifar100} in our primary experimental configuration. 
The dataset consists of 32x32 images split into 50'000 training and 10'000 test samples, evenly split between 100 classes.
For hyperparameter tuning, we use a random subset of 10\% of the training dataset as validation set.
Our reported outcomes stem from test set results associated with the most effective hyperparameters, established through retraining on the complete training set. The training duration for all models spans 200 epochs, featuring a cosine decay learning rate schedule alongside a 5-epoch warmup, with a batch size of 256.
Given the dataset's relatively modest size and the limited number of images per class, the significance of regularization is amplified.
ResNet-18 is a moderately sized network that helps keep the computational cost of training reasonable, thereby permitting multiple runs under various settings.
For all experiments, a learning rate of $\eta=0.3$ is employed. This parameter was tuned on the validation set.

\paragraph{ResNet18 + CIFAR-100}
We use the data augmentation from the original ResNet~\cite{he2016deep} paper, i.e.\ random flips and a zero padding of 2 with a random crop of the original size.
Inputs are normalized with mean \verb|(0.5071, 0.4867, 0.4408)| and std \verb|(0.2675, 0.2565, 0.2761)|.
Optimization is performed using SGD with momentum $\alpha=0.9$ using a learning rate $\eta=0.3$ and weight decay $\lambda=5 \cdot 10^{-4}$ (gains and biases excluded) at a batch size (accelerator, optimizer) of $256$.
The learning rate is tuned on the validation set, and for other hyperparameters, we use recommended default values.
A cosine decay schedule is used for the learning rate, with a 5-epoch warmup (stepwise).
We train for a total of 200 epochs (which are shorter when using a portion of the train set for validation).
One run takes around 45 minutes on a V100 GPU (baseline with batch normalization) and around 60 minutes with GNI (without any kernels or other optimizations).
We use the same setup for the LayerNorm + Weight Standardization experiments.
The Weight Standardization includes a learnable gain like in Normalization Free Networks~\cite{brock2021high} and scales the output to match the initialization magnitude.

\paragraph{ResNet20 + CIFAR-10}
The setup for CIFAR-10~\cite{cifar100} is the same as for CIFAR-100 except otherwise stated.
The inputs are normalized with mean \verb|(0.4914, 0.4822, 0.4465)| and std \verb|(0.2023, 0.1994, 0.2010)|.
The base learning rate is $\eta = 0.2$ and the weight decay $\lambda = 2\cdot 10^{-4}$.
One run takes around 14 minutes on a V100 (batch norm baseline) and 25 minutes with GNI (without any optimizations for speed).

\paragraph{ResNet50 + ImageNet-1k}
The data augmentation is based on the original ResNet~\cite{he2016deep} paper.
The training transforms consists of a torchvision RandomResizedCrop to 224x224 using the default scaling range of $[0.08, 1.0]$ and ratio range of $[0.75, 1.33]$, a random horizontal flip and a normalization for mean \verb|(0.485, 0.456, 0.406)| and std \verb|(0.229, 0.224, 0.225)|.
For inference, we resize input images to 256 and center crop 224.
The optimizer is SGD with momentum $\alpha=0.9$ using a learning rate $\eta=0.1$ and weight decay $\lambda=10^{-4}$. A batch size of $256$ (optimizer and accelerator batch size) is applied. 
We use a step-wise cosine decay learning rate schedule, with a 1 epoch warmup, and train for a total of 90 epochs.
Standard training takes around 31 hours on a single V100 GPU using float16 mixed precision.

\subsection{Extension to Other Architectures}
\paragraph{NF-ResNet + CIFAR-10} The network used is a Normalization Free ResNet proposed by \citet{brock2021high}.
We modified the existing implementation in the TIMM library~\cite{rw2019timm} to use an input block similar to the CIFAR style ResNets, i.e. a single 3x3 convolution without any stride or pooling.
The network used in the experiments has a depth 26, uses bottleneck blocks, and has 4 stages (operating on different spatial sizes) each with 2 blocks.
The number of channels in the stages is \verb|(256, 512, 1024, 2048)| between the stages and one-fourth of that in the 3x3 convolutions.
We enable the use of SkipInit in our experiments (disabled by default in TIMM) and use the default alpha value of 0.2.
When using GNI we place it right before each activation function (ReLU).
The inputs are normalized with mean \verb|(0.4914, 0.4822, 0.4465)| and std \verb|(0.2023, 0.1994, 0.2010)| like in our other CIFAR-10 experiments.
The data augmentation (when used) is a random horizontal flip with probability 0.5, padding of 4 zero values on each side followed by selecting a random 32x32 patch of the resulting image.
We train for 200 epochs using a cosine decay schedule with a 20-epoch learning rate warmup, updating the learning rate after every step.
Optimization is performed using SGD with momentum 0.9, learning rate $2 \cdot 10^{-1}$, and weight decay $5 \cdot 10^{-5}$ at a batch size of 256 in float16-float32 mixed precision.
One run takes around 50 minutes on a V100 (baseline) and 75 minutes with GNI (without speed optimizations).

\paragraph{Simple ViT + CIFAR-10} The Simple Visual Transformer is a modification of the original ViT~\cite{dosovitskiy2021image} proposed by some of the original authors in a technical report~\cite{beyer2022better}.
Our implementation is based on the ViT-PyTorch repository~\cite{vitpytorch}.
The network we train has 6 blocks (depth), 16 heads, embedding dimension of 512, MLP dimension of 1024, patch size of 4x4 (on the original input size of 32x32), and uses GELU activations.
When using GNI we insert it after every Layer Normalization (including the affine transformation).
The inputs are normalized with mean \verb|(0.4914, 0.4822, 0.4465)| and standard deviation \verb|(0.2023, 0.1994, 0.2010)| like in our other CIFAR-10 experiments.
The data augmentation (when used) is a random horizontal flip with probability 0.5, padding of 4 zero values on each side followed by selecting a random 32x32 patch of the resulting image.
We train for 200 epochs using a cosine decay schedule with a 40-epoch learning rate warmup, updating the learning rate after every step.
Optimization is performed using AdamW with learning rate $10^{-3}$, weight decay $10^{-4}$, $\beta_1=0.9$, $\beta_2=0.999$, $\epsilon=10^{-8}$ at a batch size of 256 in float16-float32 mixed precision.
One run takes around 45 minutes on a V100 (baseline) and 50 minutes with GNI (without speed optimizations).

\paragraph{ConvMixer + CIFAR-10}
\citet{} proposed the ConvMixer architecture.
We use the implementation from the TIMM library~\cite{rw2019timm}.
The network used in the experiments has a depth of 8, dimension of 256, kernel size 5 and a patch size of 2x2 (on the original input size of 32x32).
The GNI is applied inside of the batch normalization layers (between normalization and the affine transformation).
The inputs are normalized with mean \verb|(0.4914, 0.4822, 0.4465)| and std \verb|(0.2023, 0.1994, 0.2010)| like in our other CIFAR-10 experiments.
The data augmentation (when used) is a random horizontal flip with probability 0.5, padding of 4 zero values on each side followed by selecting a random 32x32 patch of the resulting image.
We train for 200 epochs using a cosine decay schedule with a 5 epoch learning rate warmup, updating the learning rate after every step.
Optimization is performed using AdamW with learning rate $2 \cdot 10^{-2}$, weight decay $5 \cdot 10^{-2}$, $\beta_1=0.9$, $\beta_2=0.999$, $\epsilon=10^{-3}$ at a batch size of 256 in float16-float32 mixed precision.
One run takes around 60 minutes on a V100 (baseline) and 80 minutes with GNI (without speed optimizations).

\section{Unbounded output range of XBN}
\label{apendix - XBN}
Below we analyze the possible output range for Ghost Batch Normalization (GBN) and Exclusive Batch Normalization (XBN), as indicated in Section~\ref{section-XBN}.
We show that in GBN the range of the output is bounded but XBN offers no such guarantees.
This could be the source of the training instability observed with XBN at smaller batch sizes.
Removing the self-dependency in the manner performed in XBN therefore results in a trade-off between a lower train-test discrepancy vs training stability.

\paragraph{Bounded output range of GBN}
Let $\vx_1,\ldots,\vx_N$ be a batch of inputs for a fully connected network (i.e. no spatial dimensions for simplicity).
Then the output of Ghost Batch Normalization corresponding to an input $\vx_i$ is given by:
\begin{equation}
\label{bounded-GBN-range}
    \frac{\vx_i - \frac{1}{N}\sum_j \vx_j}{\sqrt{\frac{1}{N}\sum_j (\vx_j - \frac{1}{N}\sum_t \vx_t)^2+\epsilon }} =  \frac{\vx_i - \frac{1}{N}\sum_j \vx_j}{\sqrt{\frac{1}{N} \left((\vx_i -\frac{1}{N}\sum_t \vx_t)^2+\sum_{j\neq i} (\vx_j - \frac{1}{N}\sum_t \vx_t)^2\right)+\epsilon }}
`\end{equation}
For analysis purposes, we treat $\epsilon$ as 0 for now. The reciprocal of \eqref{bounded-GBN-range} is:
\begin{equation}
    \frac{\sqrt{\frac{1}{N}\sum_j (\vx_j - \frac{1}{N}\sum_t \vx_t)^2}} {\vx_i - \frac{1}{N}\sum_j \vx_j} = \sqrt{\left(\frac{1}{N} +\sum_{j\neq i} \frac{(\vx_j - \frac{1}{N}\sum_t \vx_t)^2}{(\vx_i -\frac{1}{N}\sum_t \vx_t)^2}\right)} \geq \frac{1}{\sqrt{N}}
\end{equation}
which implies that the magnitude of the output is bounded by the square root of the ghost batch size.

\paragraph{Unbounded output range of XBN} 
With XBN the output for a single element $\vx_i$ will be: 
\begin{equation}
\label{unbounded-range-XBN}
\begin{aligned}
    \frac{\vx_i - \frac{1}{N-1}\sum_{j\neq i} \vx_j}{\sqrt{\frac{1}{N-1}\sum_{j\neq i} (\vx_j - \frac{1}{N-1}\sum_{t\neq i} \vx_t)^2+\epsilon}} 
\end{aligned}
\end{equation}
The numerator here has no dependency on $\vx_i$ and if we ignore $\epsilon$, it can therefore be arbitrarily small in comparison, making the output unbounded.
For example when all $\vx_j, j \ne i$ are identical, the numerator of \eqref{unbounded-range-XBN} will be $\sqrt{\epsilon}$, which is close to 0 in practice (the default value in PyTorch is $\epsilon = 10^{-5}$).
Large values like these can destabilize training.
We observe this in practice when using smaller batch sizes, due to $\vsigma_i^2$ randomly being close to zero for some ghost batches.
It may be possible to mitigate this issue, for example by clamping the computed sigma or the resulting output, but we do not pursue this further here.

\section{Derivation of \eqref{general-GNI}}\label{derivation - eq9}
Consider the double normalization setup described in Section~\ref{double normalization}, repeated below:
\begin{align}
    \hat{\Xv} &:= \highlightgreen{\text{BN}}(\mX) = \frac{\mX-\muv}{\vsigma} =[\hat{\Xv}_1,..,\hat{\mX}_G] \label{eq:apdx_bn_gbn_1}\\
    \widetilde{\Xv}_g &:= \highlightred{\text{GBN}}(\hat{\mX})_g = \frac{\hat{\mX}_g-\hat{\vmu}_g}{\hat{\vsigma}_g} \label{eq:apdx_bn_gbn_2}
\end{align}
We identified the second transformation (Equation~\ref{eq:apdx_bn_gbn_2}) as the source of the increased noise and train-test discrepancy in Ghost Batch Normalization.
Our goal in Ghost Noise Injection is twofold, decoupling the noise from the normalization and eliminating the increased train-test discrepancy.

\paragraph{Eliminating the Normalization} We first show how we can decouple Equation~\ref{eq:apdx_bn_gbn_2} from Equation~\ref{eq:apdx_bn_gbn_1}.
The output $\widetilde{\Xv}_g$ in Equation~\ref{eq:apdx_bn_gbn_2} can be written as:
\begin{align}
    \widetilde{\mX}_g &= \frac{1}{\hat{\vsigma}_g}\left(\frac{\mX_g -\vmu}{\vsigma}-\hat{\vmu}_g \right) \\
    &= \frac{\mX_g}{\hat{\vsigma}_g \vsigma} - \frac{\vmu}{\hat{\vsigma}_g \vsigma} - \frac{\hat{\vmu}_g}{\hat{\vsigma}_g}
\end{align}
We can undo the effects of the original normalization by multiplying this output by $\vsigma$ and then reverse the relative shift by adding $\vmu/\hat{\vsigma}_g$, accounting by the fact that in $\widetilde{\mX}_g$ the original shift has been scaled by $\hat{\vsigma}_g^{-1}$.
We will refer to this ``denormalized'' output of Equation~\ref{eq:apdx_bn_gbn_2} as $\text{GBN}_\Delta$ which we can write as:
\begin{align}
    \text{GBN}_\Delta(\mX) :&= \vsigma \widetilde{\mX}_g + \frac{\vmu}{\hat{\vsigma}_g} \\
    &= \frac{\mX - \vsigma \hat{\vmu}_g}{\hat{\vsigma}_g} \label{eq:apdx_gbn_delta}
\end{align}
The GBN statistics $\hat{\vmu}_g$ and $\hat{\vsigma}_g$ can be expressed in terms of the original inputs $\mX$:
\begin{align}
    \hat{\vmu}_g &= \frac{1}{NHW}\sum_{n,h,w} \hat{\mX}_g \\
    &= \frac{1}{NHW}\sum_{n,h,w} \left( \frac{\mX_g -\vmu}{\vsigma}\right) \\
    &= \frac{1}{\vsigma} \Big( \frac{1}{NHW}\sum_{n,h,w}(\mX_g) - \vmu \Big) \\
    &=: \frac{\vmu_g - \vmu}{\vsigma}
\end{align}
and we can similarly show that:
\begin{align}
    \hat{\vsigma}_g^2 &= \frac{1}{\vsigma^2} \frac{1}{NHW}\sum_{n,h,w}(\mX_g - \vmu_g)^2 \\
    &=: \vsigma_g^2/\vsigma^2
\end{align}
Plugging this into Equation~\ref{eq:apdx_gbn_delta} gives:
\begin{equation}
    \text{GBN}_\Delta (\mX) = \frac{\mX - (\vmu_g - \vmu)}{\vsigma_g / \vsigma} \label{eq:apdx_gbn_delta_final}
\end{equation}

\paragraph{Reducing the Train-Test Discrepancy}
To mitigate the increased train-test discrepancy with smaller batch sizes in GBN, we propose to re-sample ghost batches for the calculation of normalization statistics.
In Section~\ref{sec:gni} we described how we can replace Equation~\ref{eq:apdx_bn_gbn_2} with Equation~\ref{eq:double_norm_gni}, repeated below:
\begin{equation}
   \widetilde{\Xv} = \frac{\hat{\Xv}-\hat{\rvm}}{\hat{\rvs}} \label{eq:apdx_double_norm_gni}
\end{equation}
Where $\hat{\rvm}$ and $\hat{\rvs}$ are analogous to $\hat{\vmu}$ and $\hat{\vsigma}$, except they are computed over a \emph{random} subset of $\hat{\Xv}$ for each element in the batch.
By doing this, each sample only contributes weakly to its own normalization statistics in expectation, similar to batch normalization over the full accelerator batch instead of ghost batch normalization over a smaller ghost batch.
We can extend this trick to $\text{GBN}_\Delta (\mX)$ from Equation~\ref{eq:apdx_gbn_delta_final}, replacing $\vmu_g$ and $\vsigma_g$ with $\rvm$ and $\rvs$, computed like $\hat{\rvm}$ and $\hat{\rvs}$ except \emph{on the original inputs $\mX$ instead of $\hat{\mX}$}.
This gives the expression for GNI in Equation~\ref{general-GNI}:
\begin{equation}
    \text{GNI}(\mX) = \frac{\Xv-(\rvm-\muv)}{\rvs/\vsigma}
\end{equation}

\section{Distribution of ghost noise}
\label{appendix - noise distr}
This part presents the calculation details of the noise distributions derived in Section~\ref{noise distr - Conv}.
For $\hat{\mu}_{gc}$ we have:
\begin{equation}
  \E [\hat{\mu}_{gc}] = \E\left[\frac{1}{NI}\sum_n \sum_i x_{nci}\right] = \frac{1}{NI}\sum_n \sum_i \E[x_{nci}] = 0
\end{equation}
\begin{align}
    \Var[\hat{\mu}_{gc}] & = \E\left[\frac{1}{NI}\sum_n \sum_i x_{nci}\right]^2\\
    & =  \E\left[\frac{1}{NI}\sum_n \sum_i x_{nci}-\mu_{nc} +\mu_{nc}\right]^2\\
    & = \frac{1}{N^2I^2}\sum_n \sum_i \sigma_{Ic}^2 + \frac{1}{N^2I^2} NI^2 \sigma_{Bc}^2\\
    & = \frac{1}{NI}\sigma_{Ic}^2 +\frac{1}{N}\sigma_{Bc}^2
\end{align}

For $\hat{\sigma}_{gc}^2$ we have:
  \begin{align}
    \hat{\sigma}_{gc}^2 &= \frac{1}{NI} \sum_n \sum_i (x_{nci}-\mu_{nc}+\mu_{nc})^2 \\
     &= \frac{1}{NI}\sum_n \sum_i \big((x_{nci}-\mu_{nc})^2 + (\mu_{nc}-\mu)^2
     + 2(x_{nci}-\mu_{nc})(\mu_{nc}-\mu)\big) \\
     &\overset{\overline{x}_{nc}\approx \E 
     x_{nc} }{\approx} \frac{1}{NI}\sum_n \sum_i \sigma_{Ic}^2 \left(\frac{x_{nci}-\mu_{nc}}{\sigma_{Ic}}\right)^2
     + \frac{1}{N} \sigma_{Bc}^2 \sum_n \left(\frac{\mu_{nc}-\mu}{\sigma_{Bc}}\right)^2 + \frac{1}{N} (\mu_{nc}-\mu_{nc}) (\mu_{nc}-\mu)\\
    &= \frac{1}{NI}\sum_n \sum_i \sigma_{Ic}^2 \left(\frac{x_{nci}-\mu_{nc}}{\sigma_{Ic}}\right)^2 + \frac{1}{N} \sigma_{Bc}^2 \sum_n \left(\frac{\mu_{nc}-\mu}{\sigma_{Bc}}\right)^2
  \end{align}
\section{Additional Experimental Results}\label{appx:more_experiments}

\begin{figure*}[tb]
    \centering
    \begin{subfigure}[b]{0.24\textwidth}
    \includegraphics[width=\linewidth]{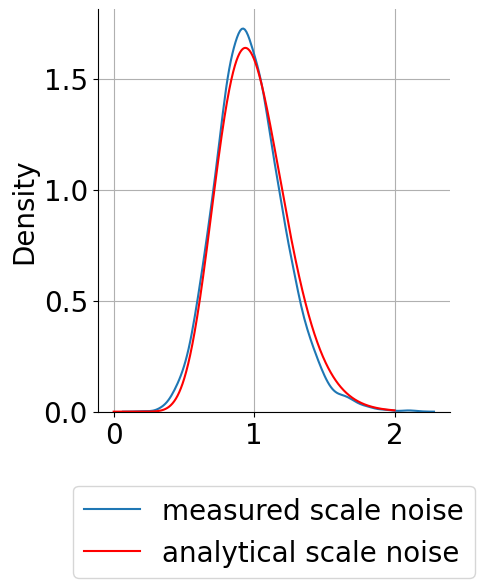}
    \centering
    \caption{Scale noise - epoch 1}
    \label{fig:fnn-scale-noise-distr1}
  \end{subfigure}
  \begin{subfigure}[b]{0.24\textwidth}
    \includegraphics[width=\linewidth]{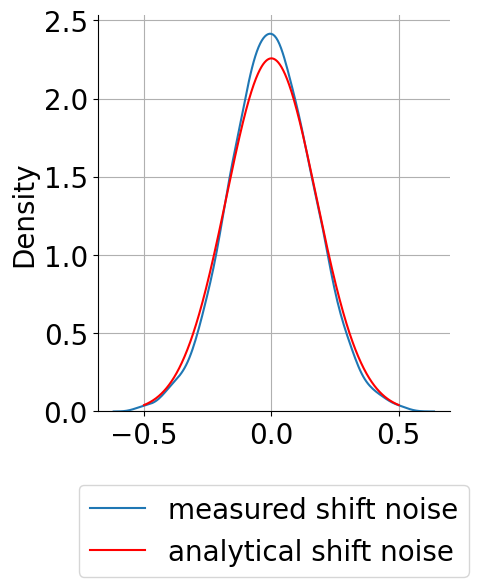}
    \caption{Shift noise - epoch 1}
    \label{fig:fnn-shift-noise-distr1}
  \end{subfigure}
  \begin{subfigure}[b]{0.24\textwidth}
    \includegraphics[width=\linewidth]{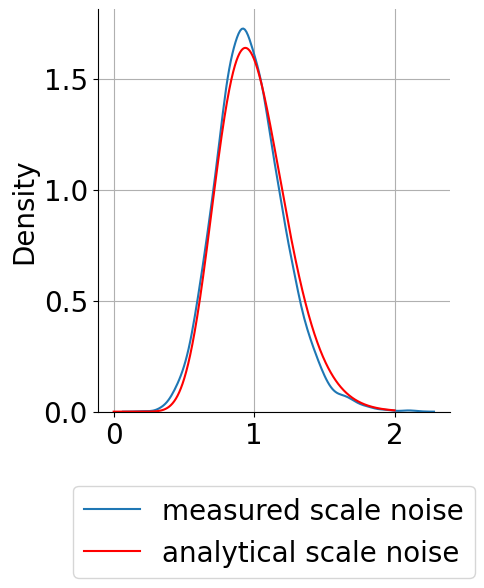}
    \caption{Scale noise - epoch 100}
    \label{fig:fnn-scale-noise-distr100}
  \end{subfigure}
  \begin{subfigure}[b]{0.24\textwidth}
    \includegraphics[width=\linewidth]{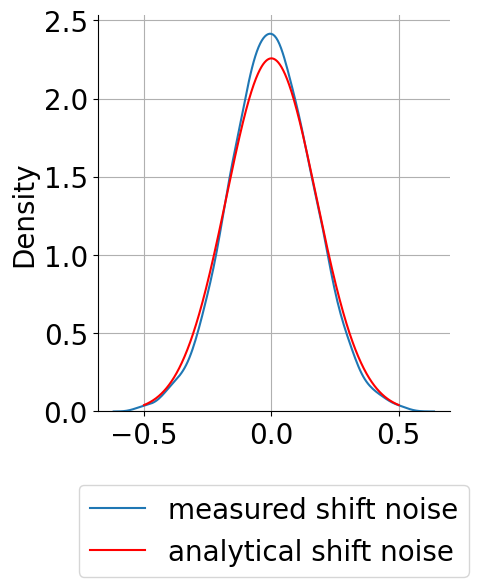}
    \caption{Shift noise - epoch 100}
    \label{fig:fnn-shift-noise-distr100}
  \end{subfigure}
  \caption{Measured scale and shift noise distributions from GNI versus our calculated analytical distribution in FNN}
  \label{fig:FNN-noise-distr}
\end{figure*}

\paragraph{Distribution of Ghost Noise in FNNs} The ghost noise distribution for convolutional networks depends on the inter-sample and intra-sample variances which make it hard to model.
Here we study the simpler 1d case, which does not have this effect and the ghost noise distribution primarily depends on the ghost batch size. %
We use a two-layer Fully connected neural network to classify MNIST~\cite{deng2012mnist}, with 512 and 300 neurons in each layer.
The inputs are flattened image tensors.
When the ghost batch size is 32, we plot out the distribution of measured scale and shift noise, with kernel density estimation to smooth the curve.
We further plot out our calculated analytical distribution curve according to Section~\ref{noise-distr-FNN}, i.e. analytical shift noise follows a normal distribution with mean 0 and std $\sqrt{1/32}$, and analytical scale noise distribution follows a Chi-square distribution with degree of freedom 32 and scale $1/32$.
We show that our analytical solution is able to match the ghost noise distribution in 1d case.

\begin{figure}[h]
\begin{center}
\includegraphics[width=0.45\columnwidth]{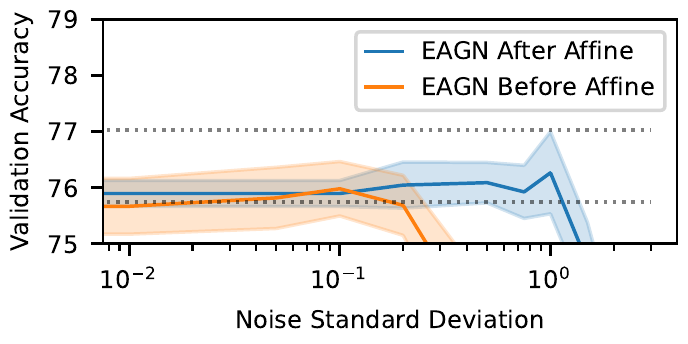}
\end{center}
\caption{Extension of Figure~\ref{fig:rn18_c100_val_overview} - Validation accuracy for EAGN}\label{fig:eagn_comparison}
\end{figure}

\paragraph{Comparison to other Noise Injection Methods}
In Section~\ref{sec:related work}, we discussed two other noise injection methods.
NoMorelization~\citep{liu2022nomorelization} is composed of two trainable scalars and a zero-centered Gaussian noise injector that is originally meant to replace the normalization layers (for use in noise-free settings).
The explicit Gaussian Noise Injections proposed by \citet{camuto2020explicit} are also additive Gaussian Noise applied after every layer except the last one.
We note that these methods are similar to AGNI-$\mu$, but the noise is sampled elementwise instead of channelwise.
In Figure~\ref{fig:eagn_comparison} we show the effect of applying i.i.d.\ Elementwise Additive Gaussian Noise (EAGN) of a fixed variance $\sigma$ (hyperparameter) in every normalization layer in the CIFAR-100 ResNet-18 setup from Figure~\ref{fig:rn18_c100_val_overview}.
It is not fully clear how best to apply this noise so we consider two placements, either before or after the learned affine transformation.
The results are shown in the figure, EAGN can provide a small accuracy boost but is unable to match GNI (which achieves over 78\% accuracy here).

\paragraph{Applicability to MLPs}
We trained an MLP (3 hidden layers of width 1024, each with batch norm + ReLU) on CIFAR-100 with a batch size of 1024, keeping the rest of the training setup the same as in our ResNet-18 experiments.
After tuning the learning rate, the results are shown in Table~\ref{tab:mlp}.
This suggests that GNI can also be beneficial in MLPs.

\begin{table}[h]
\centering
\begin{tabular}{lccccc}
    \toprule
     & Baseline    & GNI-16 & GNI-32  \\
    \midrule
    mean$\pm$std & 38.0$\pm$0.7 & 41.3$\pm$0.9 & 40.7$\pm$0.7 \\
    \bottomrule
\end{tabular}
\caption{MLP CIFAR-100 validation accuracy [\%]}\label{tab:mlp}
\end{table}

\section{Limitations}\label{appx:limitations}
Studying stochastic regularization methods can be challenging. It requires many repeated experiments to average out the random effects and reveal the true benefits of the methods.
This limits the size and number of settings we can explore.
We have focused our efforts on ResNet-18 training on CIFAR-100, which is computationally tractable for us.
Our findings can be generalized to the other settings of ResNet-50 on ImageNet, Layer-Normalized variant of ResNet-18 on CIFAR-100 and ResNet-20 on CIFAR but we are not able to perform ablations and detailed comparisons across a wider variety of settings.

The main limitations of Ghost Noise Injection are the higher computational cost and the requirement to sample from a batch.
The computational cost is similar to normalization, which is higher than that of e.g. dropout.
The computation itself is not significant in terms of flops (where convolutions / matrix multiplications dominate) but there is some run-time overhead on GPUs, mostly due to increased global memory accesses and additional kernel launches.
When used in conjunction with batch normalization, a lot of the computation can be shared.
Without efficient kernels that can merge some of the operations, the slowdown increases relative to existing, highly optimized, PyTorch operations.
The implementation we used can therefore be quite slow compared to the highly optimized native batch normalization (sometimes almost 50\% slower), but we encounter similar slowdowns for our implementation of Ghost Batch Normalization, where there is no theoretical overhead (highlighting the importance of optimized kernels).
We note that the computational cost is only present during training, not inference.
The batch dependency could potentially be mitigated by keeping track of a history of mean and variances for multiple batches, at a somewhat increased memory cost during training.
\end{document}